\documentclass{article}
\usepackage{spconf,amsmath,epsfig}
\usepackage{spconf,amsmath,epsfig}
\usepackage{amsmath}
\usepackage{amssymb}
\usepackage{upgreek}
\usepackage{amsfonts}
\usepackage{makecell}
\usepackage{array}
\usepackage{units}
\usepackage{multirow}
\usepackage{booktabs}
\usepackage{tabularx}
\usepackage{url}
\usepackage[square,numbers]{natbib}

\usepackage{tikz}
\usepackage{pgfplots}
\usepgfplotslibrary{colorbrewer}
\pgfplotsset{cycle list/Set1-9}
\tikzset{every picture/.style={line width=1pt}}
\usetikzlibrary{patterns}
\usepgfplotslibrary{fillbetween}
\usepackage[eulergreek]{sansmath}
\pgfplotsset{
  tick label style = {font=\sansmath\sffamily\scriptsize},
  every axis label = {font=\sansmath\sffamily\scriptsize},
  y label style={at={(0.05,0.5)}},
  legend style = {font=\sansmath\sffamily\scriptsize},
  label style = {font=\sansmath\sffamily\scriptsize},
}
\pgfplotsset{compat=1.3} %
\DeclareMathAlphabet\mathbfcal{OMS}{cmsy}{b}{n}

\title{A Novel Framework to Jointly Compress and Index Remote Sensing Images for Efficient Content-based Retrieval}
\name{Gencer Sumbul$^{\ast}$, Jun Xiang$^{\ast}$, Nimisha Thekke Madam, Beg{\"u}m Demir%
\thanks{$^{\ast}$ The first two authors equally contributed to the paper.}
}%
\address{Faculty of Electrical Engineering and Computer Science, Technische Universit\"at Berlin, Germany}
\begin{document}
%
\maketitle
\begin{abstract}
Remote sensing (RS) images are usually stored in compressed format to reduce the storage size of the archives. Thus, existing content-based image retrieval (CBIR) systems in RS require decoding images before applying CBIR (which is computationally demanding in the case of large-scale CBIR problems). To address this problem, in this paper, we present a joint framework that simultaneously learns RS image compression and indexing. Thus, it eliminates the need for decoding RS images before applying CBIR. The proposed framework is made up of two modules. The first module compresses RS images based on an auto-encoder architecture. The second module produces hash codes with a high discrimination capability by employing soft pairwise, bit-balancing and classification loss functions. We also introduce a two stage learning strategy with gradient manipulation techniques to obtain image representations that are compatible with both RS image indexing and compression. Experimental results show the efficacy of the proposed framework when compared to widely used approaches in RS. The code of the proposed framework is available at 
\url{https://git.tu-berlin.de/rsim/RS-JCIF}.
\end{abstract}
\vspace{-0.09in}
\section{Introduction}
\vspace{-0.1in}
Fast and accurate content-based image retrieval (CBIR) has attracted increasing attention in remote sensing (RS). Most of the CBIR methods in RS apply image search and retrieval through linear scan by comparing the image features of query image and archive images (i.e., exhaustive search). This can be computationally demanding when the image archive is large in size, and thus impractical for large-scale CBIR applications. Accordingly, hashing based indexing methods have been recently introduced in RS. These methods aim at encoding high-dimensional image features into compact binary hash codes that are indexed in a hash table. This leads to a high efficiency on search and retrieval speed. As an example, in~\cite{Roy:2020} a metric-learning based supervised deep hashing network is introduced to learn a metric space, where the features of semantically similar images are close to each other and those of dissimilar images are separated. This method applies bit-balancing and representation penalty loss functions for learning to binarize image features, while the final hash codes are obtained through thresholding. The reader is referred to \cite{Sumbul:2021} for the detailed survey of hashing-based indexing methods in RS. 

RS images are generally stored in a compressed format due to storage limitations. Recent advances on image compression show that deep learning (DL) based compression methods outperform traditional methods such as JPEG2000 by preserving the perceptual quality of images at lower bit rates \cite{Kong2021spatialSpectral}. These methods generally include a pair of encoder and decoder for feature extraction and image reconstruction, and an entropy model for bit-rate optimization. As an example, in \cite{JointMixtureGaussianAttention} residual connections with generalized divisive normalization and attention modules are utilized in encoder and decoder blocks, while context module, mixed Gaussian module and hierarchical priors are employed for the entropy model. Despite the proven success of these methods for image compression, they do not guarantee to preserve high-level semantics of images in their compressed representations~\cite{Yang:2021}. Due to this, the direct application of indexing on compressed representations may lead to inaccurate CBIR performance. Accordingly, existing RS CBIR systems require to apply decoding on compressed RS images prior to indexing. However, this is computationally expensive, and thus time demanding for operational CBIR applications on large-scale archives. 

To address this problem, in this paper, we propose a joint framework for compression and indexing of RS images. The proposed framework consists of a DL based compression module and a deep hashing based indexing module that allow us to jointly compress and index RS images. Thus, the proposed framework does not require to decompress RS images for indexing, leading to fast CBIR. For the training of the proposed framework, we introduce a two-stage learning strategy that uses gradient manipulation techniques to learn compatible image representations for both compression and indexing. 
\vspace{-0.1in}
\section{The Proposed Joint Framework for Image Compression and Indexing}
\vspace{-0.1in}
Let $\mathcal{X}=\{\textbf{X}_i\}_{i=1}^{N}$ be an archive that includes $N$ (non-compressed) images, where $\textbf{X}_i$ is the $i$th image in the archive. We assume that a training set $\mathcal{T} \subset \mathcal{X}$ is available. Each image in $\mathcal{T}$ is annotated with a set of class labels, which describes the content of the image. The set of all labels associated to $\textbf{X}_i$ is defined by a binary vector $\boldsymbol{l}_i \in \{0,1\}^{C}$, where each element indicates the presence or absence of one of $C$ classes. The proposed joint framework aims at simultaneously learning to: i) compress each image $\textbf{X}_i$ in the archive and obtain its compressed (quantized) representation $\Tilde{\textbf{Y}}_i$; and ii) index it through its hash code $\textbf{b}_i$. To this end, the proposed framework consists of: 1) a DL based compression module to learn efficient compressed representations with optimal rate-distortion trade-off; and 2) a deep hashing based indexing module to represent the complex semantic content of images by their hash codes (see Fig. 1). To effectively train the proposed joint framework, we introduce a two stage learning strategy that includes gradient manipulation techniques. Once the proposed framework is trained on $\mathcal{T}$, quantized representations can be effectively stored for all images in the archive, while the generated hash codes (which are stored in a hash table) can be utilized for efficient CBIR applications. In detail, given a query image $\textbf{X}_q$, the proposed framework obtains its hash code $\textbf{b}_q$ and retrieves semantically similar images to $\textbf{X}_q$ from the archive by analyzing the hash table. By this way, the extra step of decoding all images in the compressed archive is avoided. Only the quantized representations of retrieved images can be further decompressed for their visualization. 

The DL based compression module of the proposed framework aims to learn efficient quantized representations $\Tilde{\mathcal{Y}}$ of $\mathcal{X}$ that are entropy coded for storage, while achieving optimal rate-distortion trade-off. To achieve RS image compression, we utilize the auto-encoder architecture proposed in \cite{JointMixtureGaussianAttention} that includes an encoder $E$ and a decoder $D$. Given an input image $\textbf{X}_i$, the representation and decoded images are characterized as follows:
\begin{equation}
\textbf{Y}_i = E(\textbf{X}_i ), \enspace \Tilde{\textbf{Y}_i} = \textbf{Y}_i + U(-\frac{1}{2},\frac{1}{2}), \enspace \hat{\textbf{X}}_i = D(\Tilde{\textbf{Y}_i}),    
\label{eq:recon_x_hat}
\end{equation}
where $U$ is a uniform noise to simulate quantization during training (which is replaced by rounding function during inference) and $\hat{\textbf{X}}_i$ is the reconstructed image, which is a distorted version of the original image. To apply entropy coding, we employ a probability distribution $P$ of $\Tilde{\textbf{Y}}_i$ given its hyper-prior encoding $\textbf{Z}_i$ (i.e., $P(\Tilde{\textbf{Y}_i}|\textbf{Z}_i)$). To model $P(\Tilde{\textbf{Y}_i}|\textbf{Z}_i)$, we utilize an auto-regressive context model based on a Gaussian mixture distribution, while a non-parametric, fully factorized density model is applied to model $P(\textbf{Z}_i)$ as proposed in \cite{JointMixtureGaussianAttention}. By utilizing this probabilistic model and arithmetic coding algorithm, the quantized representation $\Tilde{\textbf{Y}_i}$ is stored in the archive using a total bit-rate from the representation and side information as $r(\Tilde{\textbf{Y}_i})+ r(\textbf{Z}_i) =-\log(P(\Tilde{\textbf{Y}_i}|\textbf{Z}_i) -\log(P(\textbf{Z}_i))$, where $P(\textbf{Z}_i)$ is modeled using a parametric entropy. To achieve a lower bit-rate with the possible minimum distortion, the compression loss function $\mathcal{L}_{C}$ is formulated as follows:
\begin{equation}
    \mathcal{L}_{C}=r(\Tilde{\textbf{Y}_i})+ r(\textbf{Z}_i)+ \lambda d(\textbf{X}_i,\hat{\textbf{X}_i}),
    \label{eq:rdtradeoff}
\end{equation}
where $\lambda$ controls the rate-distortion trade-off and $d$ defines the distortion between the original and reconstructed images.

\begin{figure}[t]
    \centering

\tikzset{every picture/.style={line width=0.75pt}} 

\begin{tikzpicture}[x=0.75pt,y=0.75pt,yscale=-1,xscale=1,scale=0.85]

\draw  [fill={rgb, 255:red, 155; green, 155; blue, 155 }  ,fill opacity=0.4 ][dash pattern={on 4.5pt off 4.5pt}] (23.1,116.47) -- (379.77,116.47) -- (379.77,430.78) -- (23.1,430.78) -- cycle ;
\draw  [fill={rgb, 255:red, 128; green, 128; blue, 128 }  ,fill opacity=0.4 ][dash pattern={on 4.5pt off 4.5pt}] (22.67,4.56) -- (379.1,4.56) -- (379.1,108.29) -- (22.67,108.29) -- cycle ;
\draw  [fill={rgb, 255:red, 255; green, 255; blue, 255 }  ,fill opacity=1 ] (33,147) -- (82.67,147) -- (82.67,167.33) -- (33,167.33) -- cycle ;
\draw    (83,156.33) -- (140,156.33) ;
\draw [shift={(143,156.33)}, rotate = 180] [fill={rgb, 255:red, 0; green, 0; blue, 0 }  ][line width=0.08]  [draw opacity=0] (7.14,-3.43) -- (0,0) -- (7.14,3.43) -- cycle    ;
\draw    (3,156.67) -- (29.8,156.25) ;
\draw [shift={(32.8,156.2)}, rotate = 179.1] [fill={rgb, 255:red, 0; green, 0; blue, 0 }  ][line width=0.08]  [draw opacity=0] (7.14,-3.43) -- (0,0) -- (7.14,3.43) -- cycle    ;
\draw  [fill={rgb, 255:red, 255; green, 255; blue, 255 }  ,fill opacity=1 ] (75.67,186.93) -- (148.67,186.93) -- (148.67,207.27) -- (75.67,207.27) -- cycle ;
\draw    (113,156.33) -- (113,182.33) ;
\draw [shift={(113,185.33)}, rotate = 270] [fill={rgb, 255:red, 0; green, 0; blue, 0 }  ][line width=0.08]  [draw opacity=0] (7.14,-3.43) -- (0,0) -- (7.14,3.43) -- cycle    ;
\draw  [fill={rgb, 255:red, 255; green, 255; blue, 255 }  ,fill opacity=1 ] (144.67,140.67) -- (195.67,140.67) -- (195.67,176.67) -- (144.67,176.67) -- cycle ;
\draw    (112.67,208) -- (113,227.67) ;
\draw  [fill={rgb, 255:red, 255; green, 255; blue, 255 }  ,fill opacity=1 ] (30.33,243.67) -- (90.67,243.67) -- (90.67,279.67) -- (30.33,279.67) -- cycle ;
\draw  [fill={rgb, 255:red, 255; green, 255; blue, 255 }  ,fill opacity=1 ] (139.67,242.67) -- (187.67,242.67) -- (187.67,278.67) -- (139.67,278.67) -- cycle ;
\draw    (62.67,239.33) -- (62.67,227.67) -- (163.33,227.67) -- (163.07,238) ;
\draw [shift={(163,241)}, rotate = 271.43] [fill={rgb, 255:red, 0; green, 0; blue, 0 }  ][line width=0.08]  [draw opacity=0] (7.14,-3.43) -- (0,0) -- (7.14,3.43) -- cycle    ;
\draw [shift={(62.67,242.33)}, rotate = 270] [fill={rgb, 255:red, 0; green, 0; blue, 0 }  ][line width=0.08]  [draw opacity=0] (7.14,-3.43) -- (0,0) -- (7.14,3.43) -- cycle    ;
\draw  [fill={rgb, 255:red, 0; green, 0; blue, 0 }  ,fill opacity=1 ] (29.67,306.67) -- (40,306.67) -- (40,317) -- (29.67,317) -- cycle ;
\draw  [fill={rgb, 255:red, 0; green, 0; blue, 0 }  ,fill opacity=1 ] (71,306.67) -- (81.33,306.67) -- (81.33,317) -- (71,317) -- cycle ;
\draw  [fill={rgb, 255:red, 0; green, 0; blue, 0 }  ,fill opacity=1 ] (50.33,306.67) -- (60.67,306.67) -- (60.67,317) -- (50.33,317) -- cycle ;
\draw  [fill={rgb, 255:red, 255; green, 255; blue, 255 }  ,fill opacity=1 ] (40,306.67) -- (50.33,306.67) -- (50.33,317) -- (40,317) -- cycle ;
\draw  [fill={rgb, 255:red, 255; green, 255; blue, 255 }  ,fill opacity=1 ] (60.67,306.67) -- (71,306.67) -- (71,317) -- (60.67,317) -- cycle ;
\draw  [fill={rgb, 255:red, 255; green, 255; blue, 255 }  ,fill opacity=1 ] (81.33,306.67) -- (91.67,306.67) -- (91.67,317) -- (81.33,317) -- cycle ;
\draw    (60.33,280) -- (60.63,303.67) ;
\draw [shift={(60.67,306.67)}, rotate = 269.28] [fill={rgb, 255:red, 0; green, 0; blue, 0 }  ][line width=0.08]  [draw opacity=0] (7.14,-3.43) -- (0,0) -- (7.14,3.43) -- cycle    ;
\draw  [fill={rgb, 255:red, 255; green, 255; blue, 255 }  ,fill opacity=1 ] (29.67,345) -- (90,345) -- (90,381) -- (29.67,381) -- cycle ;
\draw    (60.67,317) -- (60.96,340.67) ;
\draw [shift={(61,343.67)}, rotate = 269.28] [fill={rgb, 255:red, 0; green, 0; blue, 0 }  ][line width=0.08]  [draw opacity=0] (7.14,-3.43) -- (0,0) -- (7.14,3.43) -- cycle    ;
\draw  [fill={rgb, 255:red, 255; green, 255; blue, 255 }  ,fill opacity=1 ] (33.67,405.93) -- (83.33,405.93) -- (83.33,426.27) -- (33.67,426.27) -- cycle ;
\draw    (6.67,415.55) -- (33.47,415.13) ;
\draw [shift={(3.67,415.6)}, rotate = 359.1] [fill={rgb, 255:red, 0; green, 0; blue, 0 }  ][line width=0.08]  [draw opacity=0] (7.14,-3.43) -- (0,0) -- (7.14,3.43) -- cycle    ;
\draw    (60,381) -- (60,401.67) ;
\draw [shift={(60,404.67)}, rotate = 270] [fill={rgb, 255:red, 0; green, 0; blue, 0 }  ][line width=0.08]  [draw opacity=0] (7.14,-3.43) -- (0,0) -- (7.14,3.43) -- cycle    ;
\draw  [fill={rgb, 255:red, 255; green, 255; blue, 255 }  ,fill opacity=1 ] (255,147.93) -- (328,147.93) -- (328,168.27) -- (255,168.27) -- cycle ;
\draw  [fill={rgb, 255:red, 255; green, 255; blue, 255 }  ,fill opacity=1 ] (208.67,205) -- (274.33,205) -- (274.33,255.67) -- (208.67,255.67) -- cycle ;
\draw  [fill={rgb, 255:red, 255; green, 255; blue, 255 }  ,fill opacity=1 ] (310.67,210.67) -- (371,210.67) -- (371,246.67) -- (310.67,246.67) -- cycle ;
\draw    (241.04,201.67) -- (241.22,189.89) -- (342.89,190.89) -- (342.98,207.33) ;
\draw [shift={(343,210.33)}, rotate = 269.67] [fill={rgb, 255:red, 0; green, 0; blue, 0 }  ][line width=0.08]  [draw opacity=0] (7.14,-3.43) -- (0,0) -- (7.14,3.43) -- cycle    ;
\draw [shift={(241,204.67)}, rotate = 270.86] [fill={rgb, 255:red, 0; green, 0; blue, 0 }  ][line width=0.08]  [draw opacity=0] (7.14,-3.43) -- (0,0) -- (7.14,3.43) -- cycle    ;
\draw    (292,168.67) -- (292.05,190.39) ;
\draw    (273.86,228.86) -- (307.19,228.86) ;
\draw [shift={(310.19,228.86)}, rotate = 180] [fill={rgb, 255:red, 0; green, 0; blue, 0 }  ][line width=0.08]  [draw opacity=0] (7.14,-3.43) -- (0,0) -- (7.14,3.43) -- cycle    ;
\draw  [fill={rgb, 255:red, 0; green, 0; blue, 0 }  ,fill opacity=1 ] (310.33,273.67) -- (320.67,273.67) -- (320.67,284) -- (310.33,284) -- cycle ;
\draw  [fill={rgb, 255:red, 0; green, 0; blue, 0 }  ,fill opacity=1 ] (351.67,273.67) -- (362,273.67) -- (362,284) -- (351.67,284) -- cycle ;
\draw  [fill={rgb, 255:red, 0; green, 0; blue, 0 }  ,fill opacity=1 ] (331,273.67) -- (341.33,273.67) -- (341.33,284) -- (331,284) -- cycle ;
\draw  [fill={rgb, 255:red, 255; green, 255; blue, 255 }  ,fill opacity=1 ] (320.67,273.67) -- (331,273.67) -- (331,284) -- (320.67,284) -- cycle ;
\draw  [fill={rgb, 255:red, 255; green, 255; blue, 255 }  ,fill opacity=1 ] (341.33,273.67) -- (351.67,273.67) -- (351.67,284) -- (341.33,284) -- cycle ;
\draw  [fill={rgb, 255:red, 255; green, 255; blue, 255 }  ,fill opacity=1 ] (362,273.67) -- (372.33,273.67) -- (372.33,284) -- (362,284) -- cycle ;
\draw    (341,247) -- (341.3,270.67) ;
\draw [shift={(341.33,273.67)}, rotate = 269.28] [fill={rgb, 255:red, 0; green, 0; blue, 0 }  ][line width=0.08]  [draw opacity=0] (7.14,-3.43) -- (0,0) -- (7.14,3.43) -- cycle    ;
\draw  [fill={rgb, 255:red, 255; green, 255; blue, 255 }  ,fill opacity=1 ] (310.33,312) -- (370.67,312) -- (370.67,348) -- (310.33,348) -- cycle ;
\draw    (341.33,284) -- (341.63,307.67) ;
\draw [shift={(341.67,310.67)}, rotate = 269.28] [fill={rgb, 255:red, 0; green, 0; blue, 0 }  ][line width=0.08]  [draw opacity=0] (7.14,-3.43) -- (0,0) -- (7.14,3.43) -- cycle    ;
\draw    (239.4,256.53) -- (239.4,329.53) -- (307.4,329.44) ;
\draw [shift={(310.4,329.44)}, rotate = 179.93] [fill={rgb, 255:red, 0; green, 0; blue, 0 }  ][line width=0.08]  [draw opacity=0] (7.14,-3.43) -- (0,0) -- (7.14,3.43) -- cycle    ;
\draw  [fill={rgb, 255:red, 255; green, 255; blue, 255 }  ,fill opacity=1 ] (130,338) -- (198.74,338) -- (198.74,391.11) -- (130,391.11) -- cycle ;
\draw  [fill={rgb, 255:red, 255; green, 255; blue, 255 }  ,fill opacity=1 ] (316.67,377) -- (368.37,377) -- (368.37,412.11) -- (316.67,412.11) -- cycle ;
\draw    (162.74,394.44) -- (163.07,420.74) -- (341.07,421.74) -- (341.07,412.41) ;
\draw [shift={(162.71,391.44)}, rotate = 89.29] [fill={rgb, 255:red, 0; green, 0; blue, 0 }  ][line width=0.08]  [draw opacity=0] (7.14,-3.43) -- (0,0) -- (7.14,3.43) -- cycle    ;
\draw    (163.67,279.67) -- (163.27,334.77) ;
\draw [shift={(163.25,337.77)}, rotate = 270.41] [fill={rgb, 255:red, 0; green, 0; blue, 0 }  ][line width=0.08]  [draw opacity=0] (7.14,-3.43) -- (0,0) -- (7.14,3.43) -- cycle    ;
\draw    (94.25,259.17) -- (106.25,259.17) -- (106.58,359.5) -- (128.92,359.5) ;
\draw [shift={(91.25,259.17)}, rotate = 0] [fill={rgb, 255:red, 0; green, 0; blue, 0 }  ][line width=0.08]  [draw opacity=0] (7.14,-3.43) -- (0,0) -- (7.14,3.43) -- cycle    ;
\draw    (106.58,359.5) -- (93.58,359.77) ;
\draw [shift={(90.58,359.83)}, rotate = 358.81] [fill={rgb, 255:red, 0; green, 0; blue, 0 }  ][line width=0.08]  [draw opacity=0] (7.14,-3.43) -- (0,0) -- (7.14,3.43) -- cycle    ;
\draw    (93.58,375.5) -- (117.25,375.5) -- (117.25,259.17) -- (139.58,258.83) ;
\draw [shift={(90.58,375.5)}, rotate = 0] [fill={rgb, 255:red, 0; green, 0; blue, 0 }  ][line width=0.08]  [draw opacity=0] (7.14,-3.43) -- (0,0) -- (7.14,3.43) -- cycle    ;
\draw    (341.67,348.33) -- (341.96,372) ;
\draw [shift={(342,375)}, rotate = 269.28] [fill={rgb, 255:red, 0; green, 0; blue, 0 }  ][line width=0.08]  [draw opacity=0] (7.14,-3.43) -- (0,0) -- (7.14,3.43) -- cycle    ;
\draw  [fill={rgb, 255:red, 255; green, 255; blue, 255 }  ,fill opacity=1 ] (86.67,48.27) -- (140.1,48.27) -- (140.1,68.02) -- (86.67,68.02) -- cycle ;
\draw  [fill={rgb, 255:red, 255; green, 255; blue, 255 }  ,fill opacity=1 ] (177.67,39.67) -- (231.77,39.67) -- (231.77,75.35) -- (177.67,75.35) -- cycle ;
\draw (113,156.33) -- (113.24,71.23) ;
\draw [shift={(113.25,68.23)}, rotate = 90.16] [fill={rgb, 255:red, 0; green, 0; blue, 0 }  ][line width=0.08]  [draw opacity=0] (7.14,-3.43) -- (0,0) -- (7.14,3.43) -- cycle    ;
\draw    (196.33,158.33) -- (250.77,158.35) ;
\draw [shift={(253.77,158.35)}, rotate = 180.02] [fill={rgb, 255:red, 0; green, 0; blue, 0 }  ][line width=0.08]  [draw opacity=0] (7.14,-3.43) -- (0,0) -- (7.14,3.43) -- cycle    ;
\draw  [fill={rgb, 255:red, 255; green, 255; blue, 255 }  ,fill opacity=1 ] (287.17,10) -- (341.27,10) -- (341.27,45.69) -- (287.17,45.69) -- cycle ;
\draw  [fill={rgb, 255:red, 255; green, 255; blue, 255 }  ,fill opacity=1 ] (273.33,67) -- (353.92,67) -- (353.92,102.69) -- (273.33,102.69) -- cycle ;
\draw    (140.52,57.2) -- (173.86,57.2) ;
\draw [shift={(176.86,57.2)}, rotate = 180] [fill={rgb, 255:red, 0; green, 0; blue, 0 }  ][line width=0.08]  [draw opacity=0] (7.14,-3.43) -- (0,0) -- (7.14,3.43) -- cycle    ;
\draw    (268.92,83.98) -- (255.69,83.84) -- (256.19,27.84) -- (284.13,28.13) ;
\draw [shift={(287.13,28.17)}, rotate = 180.61] [fill={rgb, 255:red, 0; green, 0; blue, 0 }  ][line width=0.08]  [draw opacity=0] (7.14,-3.43) -- (0,0) -- (7.14,3.43) -- cycle    ;
\draw [shift={(271.92,84.02)}, rotate = 180.65] [fill={rgb, 255:red, 0; green, 0; blue, 0 }  ][line width=0.08]  [draw opacity=0] (7.14,-3.43) -- (0,0) -- (7.14,3.43) -- cycle    ;
\draw    (354.33,83.83) -- (388.37,83.83) ;
\draw [shift={(391.37,83.83)}, rotate = 180] [fill={rgb, 255:red, 0; green, 0; blue, 0 }  ][line width=0.08]  [draw opacity=0] (7.14,-3.43) -- (0,0) -- (7.14,3.43) -- cycle    ;
\draw    (341.84,27.33) -- (388.1,26.95) ;
\draw [shift={(391.1,26.93)}, rotate = 179.53] [fill={rgb, 255:red, 0; green, 0; blue, 0 }  ][line width=0.08]  [draw opacity=0] (7.14,-3.43) -- (0,0) -- (7.14,3.43) -- cycle    ;
\draw    (231.59,57.56) -- (256.19,57.5) ;

\draw (57.05,157.8) node  [font=\footnotesize] [align=left] {\begin{minipage}[lt]{33.07pt}\setlength\topsep{0pt}
\begin{center}
Encoder
\end{center}

\end{minipage}};
\draw (74.,189.93) node [anchor=north west][inner sep=0.75pt]  [font=\footnotesize] [align=left] {\begin{minipage}[lt]{48.48pt}\setlength\topsep{0pt}
\begin{center}
Quantization
\end{center}

\end{minipage}};
\draw (143.,143.67) node [anchor=north west][inner sep=0.75pt]  [font=\footnotesize] [align=left] {\begin{minipage}[lt]{33.07pt}\setlength\topsep{0pt}
\begin{center}
Hyper \\Encoder
\end{center}

\end{minipage}};
\draw (29.,246.67) node [anchor=north west][inner sep=0.75pt]  [font=\footnotesize] [align=left] {\begin{minipage}[lt]{38.95pt}\setlength\topsep{0pt}
\begin{center}
Arithmetic\\Coder
\end{center}

\end{minipage}};
\draw (138.,245.67) node [anchor=north west][inner sep=0.75pt]  [font=\footnotesize] [align=left] {\begin{minipage}[lt]{30.8pt}\setlength\topsep{0pt}
\begin{center}
Context\\Model
\end{center}

\end{minipage}};
\draw (28.,348) node [anchor=north west][inner sep=0.75pt]  [font=\footnotesize] [align=left] {\begin{minipage}[lt]{38.95pt}\setlength\topsep{0pt}
\begin{center}
Arithmetic\\Decoder
\end{center}

\end{minipage}};
\draw (31.,410.33) node [anchor=north west][inner sep=0.75pt]  [font=\footnotesize] [align=left] {\begin{minipage}[lt]{33.53pt}\setlength\topsep{0pt}
\begin{center}
Decoder
\end{center}

\end{minipage}};
\draw (254,150.93) node [anchor=north west][inner sep=0.75pt]  [font=\footnotesize] [align=left] {\begin{minipage}[lt]{48.48pt}\setlength\topsep{0pt}
\begin{center}
Quantization
\end{center}

\end{minipage}};
\draw (207.,208) node [anchor=north west][inner sep=0.75pt]  [font=\footnotesize] [align=left] {\begin{minipage}[lt]{43.03pt}\setlength\topsep{0pt}
\begin{center}
Factorized \\Entropy\\Model
\end{center}

\end{minipage}};
\draw (309.,213.67) node [anchor=north west][inner sep=0.75pt]  [font=\footnotesize] [align=left] {\begin{minipage}[lt]{38.95pt}\setlength\topsep{0pt}
\begin{center}
Arithmetic\\Coder
\end{center}

\end{minipage}};
\draw (309.,315) node [anchor=north west][inner sep=0.75pt]  [font=\footnotesize] [align=left] {\begin{minipage}[lt]{38.95pt}\setlength\topsep{0pt}
\begin{center}
Arithmetic\\Decoder
\end{center}

\end{minipage}};
\draw (129,341) node [anchor=north west][inner sep=0.75pt]  [font=\footnotesize] [align=left] {\begin{minipage}[lt]{44.85pt}\setlength\topsep{0pt}
\begin{center}
Entropy\\Parameters\\Estimation
\end{center}

\end{minipage}};
\draw (315.,380) node [anchor=north west][inner sep=0.75pt]  [font=\footnotesize] [align=left] {\begin{minipage}[lt]{33.53pt}\setlength\topsep{0pt}
\begin{center}
Hyper\\Decoder
\end{center}

\end{minipage}};
\draw (165.9,119.33) node [anchor=north west][inner sep=0.75pt]   [align=left] {DL Based Compression Module};
\draw (85,51.27) node [anchor=north west][inner sep=0.75pt]  [font=\footnotesize] [align=left] {\begin{minipage}[lt]{34.89pt}\setlength\topsep{0pt}
\begin{center}
Attention
\end{center}

\end{minipage}};
\draw (176.,42.67) node [anchor=north west][inner sep=0.75pt]  [font=\footnotesize] [align=left] {\begin{minipage}[lt]{34.89pt}\setlength\topsep{0pt}
\begin{center}
Hashing \\Network
\end{center}

\end{minipage}};
\draw (286.,13) node [anchor=north west][inner sep=0.75pt]  [font=\footnotesize] [align=left] {\begin{minipage}[lt]{34.89pt}\setlength\topsep{0pt}
\begin{center}
Hashing \\Layer
\end{center}

\end{minipage}};
\draw (272.,70) node [anchor=north west][inner sep=0.75pt]  [font=\footnotesize] [align=left] {\begin{minipage}[lt]{53pt}\setlength\topsep{0pt}
\begin{center}
Classification \\Layer
\end{center}

\end{minipage}};
\draw (30.33,6.83) node [anchor=north west][inner sep=0.75pt]   [align=left] {Deep Hashing Based Indexing Module};
\draw (4,143.22) node [anchor=north west][inner sep=0.75pt]  [font=\scriptsize]  {$\mathbf{X}_{i}$};
\draw (4.5,394.22) node [anchor=north west][inner sep=0.75pt]  [font=\scriptsize]  {$\hat{\mathbf{X}}_{i}$};
\draw (89.5,142.72) node [anchor=north west][inner sep=0.75pt]  [font=\scriptsize]  {$\mathbf{Y}_{i}$};
\draw (214,144.22) node [anchor=north west][inner sep=0.75pt]  [font=\scriptsize]  {$\mathbf{Z}_{i}$};
\draw (294,172.07) node [anchor=north west][inner sep=0.75pt]  [font=\scriptsize]  {$\tilde{\mathbf{Z}}_{i}$};
\draw (113.33,209.07) node [anchor=north west][inner sep=0.75pt]  [font=\scriptsize]  {$\tilde{\mathbf{Y}}_{i}$};
\draw (358.17,14.72) node [anchor=north west][inner sep=0.75pt]  [font=\scriptsize]  {$\mathbf{b}_{i}$};
\draw (362.17,64.4) node [anchor=north west][inner sep=0.75pt]  [font=\scriptsize]  {$\tilde{\boldsymbol{l}}_{i}$};
\end{tikzpicture}
    \caption{Illustration of the proposed joint framework.}
\label{fig:model}
  \vspace{-0.3cm}
\end{figure}
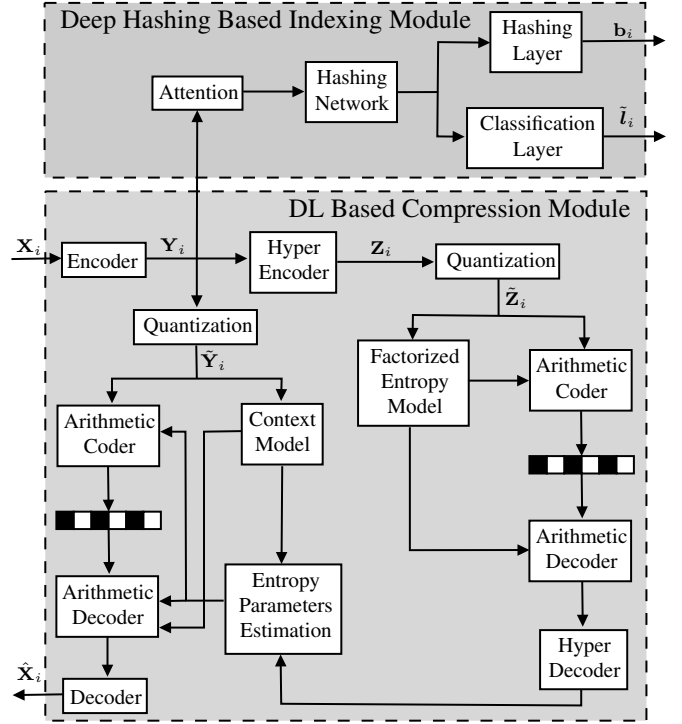
The deep hashing based indexing module of the proposed framework aims to map the encoded representations of the compression module to $\textit{q}$-bit hash codes that preserves the semantic similarities of images. Then, a hash table can be produced, where semantically similar images are located in the same hash bucket. Accordingly, indexing of the archive images is succeeded. To this end, this module includes an attention layer followed by a hashing network, which is preceded by a classification layer and a final hashing layer. For learning the hash codes, we employ the soft pairwise loss function \cite{SoftPairwise} that considers the rank difference of semantic pairwise similarities as follows:
\begin{equation}
\begin{aligned}
\mathcal{L}_{p} & = \sum_{(i,j)\in \xi}m_{ij}\Big (\log(1 + e^{\alpha s_{ij}^h}) - \alpha s_{ij}^h{s}_{ij}^o\Big )\\ &+ \gamma (1-m_{ij})\Big ( \big\lVert \frac{1}{2}(s_{ij}^h +q) - s_{ij}^o q \big\rVert^2_2\Big)  ,\\
s_{ij}^o & = \frac{<\textit{\textbf{l}}_i, \textit{\textbf{l}}_j>}{\lVert \textit{\textbf{l}}_i \rVert_2 \lVert \textit{\textbf{l}}_j \rVert_2}, \quad s_{ij}^h = <\textbf{b}_i, \textbf{b}_j>,\\
\end{aligned}
\end{equation}
where $\xi = \{(i,j)|\hspace{1mm} i, j \in \{1, ... N\}\}$ is the set of index pairs. $s_{ij}^o$ and $s_{ij}^h$ are the semantic pairwise similarities of $\textbf{X}_i$, $\textbf{X}_j$ and $\textbf{b}_i$, $\textbf{b}_j$, respectively. $m_{ij}$ is an indicator for hard similarity (i.e., $s_{ij}^{o} \in \{0,1\}$), $\alpha$ is the continuous relaxation constraint and $\gamma$ is a weighting parameter. To achieve a balanced distribution of hash codes by maximizing hash code variance, we utilize the bit-balancing loss function \cite{Fang:2018} as follows: 
\begin{equation}
    \mathcal{L}_{b}  = \sum_{(i,j)\in \xi}\Big ( \lVert (\textbf{b}_i^T\textbf{1}) \rVert^2_2 +\lVert (\textbf{b}_j^T\textbf{1}) \rVert^2_2 \Big ).
    \label{eq:balance_loss}
\end{equation}
To make the hidden features more discriminative by taking full advantage of image labels, we formulate the classification loss over image pairs as follows:
\begin{equation}
    \mathcal{L}_{c} = \sum_{(i,j) \in \xi} \Big (\lVert \hat{\textit{\textbf{l}}_i} - \textit{\textbf{l}}_i  \rVert^2_2 + \lVert \hat{\textit{\textbf{l}}_j} - \textit{\textbf{l}}_j \rVert^2_2\Big ),
    \label{eq:classification_loss}
\end{equation}
where $\hat{\textit{\textbf{l}}}_i$ and $\hat{\textit{\textbf{l}}}_j$ are predicted multi-hot label vectors. By considering the above-mentioned loss functions, the final hashing loss function is formulated as the combination of them ($\mathcal{L}_{H} = \mathcal{L}_{p} + \mathcal{L}_{b} + \mathcal{L}_{c}$).

\begin{table}[t] 
\renewcommand{\arraystretch}{0.9}
\setlength\tabcolsep{5.1pt}
\centering
\caption{Retrieval time (in seconds) and accuracies obtained by the standard approach (which requires image decoding) and our hashing based indexing module.}
\label{table:map}
\vspace{5pt} 
\begin{tabular}{
@{}*{1}{>{\raggedright\arraybackslash}p{0.169\textwidth}}*{2}{>{\centering\arraybackslash}p{0.05\textwidth}}*{1}{>{\centering\arraybackslash}p{0.075\textwidth}}*{1}{>{\centering\arraybackslash}p{0.055\textwidth}}@{}}
\toprule
Approach &$P$ (\%) & $R$ (\%) &mAP (\%) & Time \tabularnewline
\toprule
Standard approach & 72.0 & 72.0 & 71.7 & 1740.3 \tabularnewline \cmidrule{1-5} 
Our indexing module & 71.5 & 69.0 & 71.5 & 439.7 \tabularnewline 
\bottomrule
\end{tabular}
\vspace{-0.3cm}
\end{table}
It is worth emphasizing that learning image compression enforces encoded representations to retain maximum information necessary for restoring the image, while hashing based indexing enforces representations to be discriminative among images in terms of semantic image content. To make the representations compatible with both, we introduce a two stage learning strategy. In the first stage, the DL based compression module is trained alone until its convergence. Since rate and distortion in (\ref{eq:rdtradeoff}) are conflicting, we utilize the multiple-gradient descent algorithm \cite{MGDA} to obtain the optimum rate-distortion trade-off points. In the second stage, the deep hashing based indexing module is added to the compression module. To avoid the interference of different loss functions included in the hashing loss $\mathcal{L}_{H}$, we utilize PCGrad gradient manipulation technique presented in \cite{PCGrad}. We also update the compression module with a small learning rate in the second stage to make the representations compatible with indexing. 
\vspace{-0.05in}
\section{Experimental Results}
\vspace{-0.05in}
Experiments were conducted on the BigEarthNet-S2 benchmark archive \cite{BigEarthNet19}. Each image is associated with multi-labels from the 2018 CORINE Land Cover database. In this paper, we utilized the multi-labels based on the nomenclature of 19 classes ~\cite{BigEarthNet19}. To perform experiments, we selected its subset including 14,832 images acquired over Serbia on the summer season. Then, we divided this subset into training (52\%), validation (24\%) and test (24\%) sets. The query images for CBIR were selected from the validation set (3563 images were used as query), while images to be retrieved were selected from the test set. The network parameters of the DL based compression module were adapted from \cite{JointMixtureGaussianAttention}. The hashing network within the deep hashing based indexing module consists of two convolutional layers, each of which includes 512 hidden units with ReLU activation. The classification layer and the hashing layer both include single convolutional layer with the filter size of 1 followed by sigmoid and Greedy hash \cite{hashSGN} activations, respectively. The parameters of $\alpha$ and $\gamma$ were set to $5/q$ and $0.1/q$, respectively, while the hash code length $q$ was set to 64. We compared the performance of each module with the standard approaches. In detail, the results of our DL based compression module was compared with those obtained by JPEG2000. We compared the results of our deep hashing based indexing module with those obtained by an approach that uses the same hashing method applied to fully decoded images (denoted as standard approach). Experimental results are provided in terms of peak signal-to-noise ratio (PSNR) for comparing compression performances. Precision ($P$), recall ($R$), mean average precision (mAP) and retrieval time are used for comparing retrieval performances. 
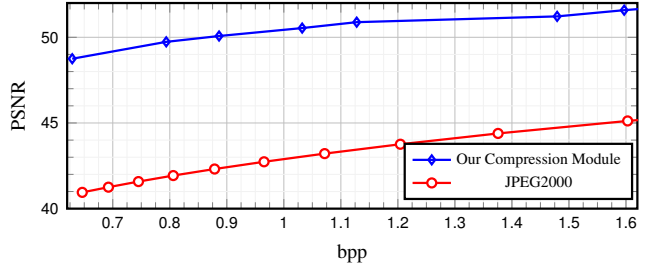
\begin{figure}[t]
    \centering
    \begin{tikzpicture}[scale = 0.85]%
	\begin{axis}[
    legend columns=1,
    height=4.8cm,
    width=10.5cm,
    legend style={font=\scriptsize, at={(axis cs:1.21,40.5)},anchor=south west},
    grid=both,
    grid style={line width=.1pt, draw=gray!10},
    major grid style={line width=.2pt,draw=gray!50},
    minor x tick num=3,
    minor y tick num=4,
    xlabel= {\small bpp},
    ylabel= {\small PSNR},
    xmin=0.62,xmax=1.62,
    ymin=40,ymax=52]
    \addplot+[name path=capacity,mark=diamond,color=blue,mark options={fill=white},line width=1pt] table [x=bpp, y=psnr, col sep=comma] {data/compression_cnn.csv};\addlegendentry{Our Compression Module};
    \addplot+[name path=capacity,color=red,mark=*,mark options={fill=white},line width=1pt] table [x=bpp, y=psnr, col sep=comma] {data/compression_jpeg2000.csv};\addlegendentry{JPEG2000};
	\end{axis}
    \end{tikzpicture}%
    \vspace{-0.4cm}
    \caption{Peak signal-to-noise ratio (PSNR) versus bits per pixel (bpp) obtained by our compression module and JPEG2000.}
    \label{fig:compression}
    \vspace{-0.4cm}
\end{figure}
\begin{figure*}[ht!]
    \newcommand{\figwidth}{0.11\linewidth}
    \newcommand{\figheight}{0.75in}
    \renewcommand{\fboxsep}{0pt}
    \centering
     \begin{minipage}[t]{\figwidth}
        \centering
        \vspace{0.45in}
        \centerline{}\medskip
        \centerline{\includegraphics[height=\figheight]{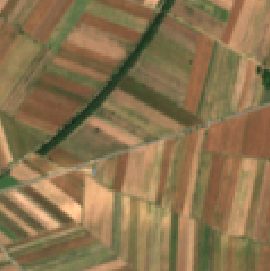}}
\vspace{-0.05in}\centerline{(a)}\medskip\vspace{-0.05in}
    \end{minipage}
     \begin{minipage}[t]{\figwidth}
        \centering
        \centerline{1\textsuperscript{st}}\medskip
        \centerline{\includegraphics[height=\figheight]{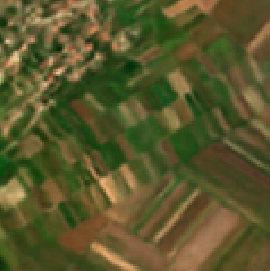}} 
    \end{minipage}
          \begin{minipage}[t]{\figwidth}
        \centering
        \centerline{2\textsuperscript{nd}}\medskip
        \centerline{\includegraphics[height=\figheight]{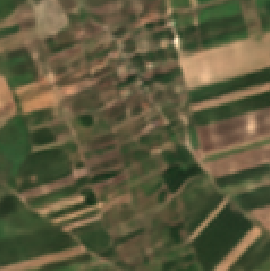}} 
    \end{minipage}    
          \begin{minipage}[t]{\figwidth}
        \centering
        \centerline{3\textsuperscript{rd}}\medskip
        \centerline{\includegraphics[height=\figheight]{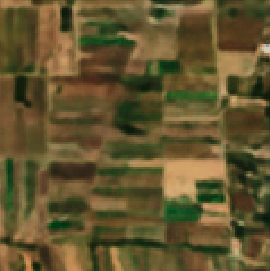}} 
    \end{minipage}
          \begin{minipage}[t]{\figwidth}
        \centering
        \centerline{4\textsuperscript{th}}\medskip
        \centerline{\includegraphics[height=\figheight]{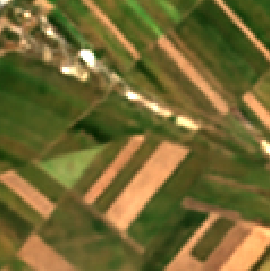}} 
        \vspace{-0.05in}\centerline{(b)}\medskip\vspace{-0.05in}
    \end{minipage}
          \begin{minipage}[t]{\figwidth}
        \centering
        \centerline{5\textsuperscript{th}}\medskip
        \centerline{\includegraphics[height=\figheight]{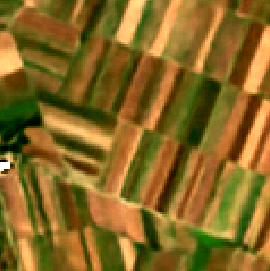}} 
    \end{minipage}
\begin{minipage}[t]{\figwidth}
        \centering
        \centerline{6\textsuperscript{th}}\medskip
        \centerline{\includegraphics[height=\figheight]{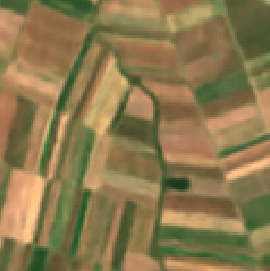}} 
    \end{minipage}
     \begin{minipage}[t]{\figwidth}
        \centering
        \centerline{7\textsuperscript{th}}\medskip
        \centerline{\includegraphics[height=\figheight]{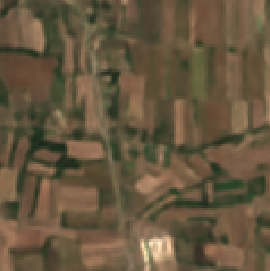}} 
    \end{minipage}
     \begin{minipage}[t]{\figwidth}
        \centering
        \vspace{-0.47in}
        \centerline{\includegraphics[height=\figheight]{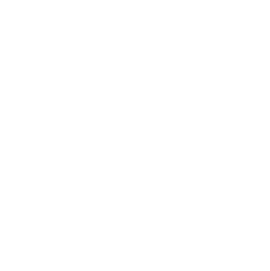}}
    \end{minipage}
          \begin{minipage}[t]{\figwidth}
        \centering
        \vspace{-0.47in}
        \centerline{\includegraphics[height=\figheight]{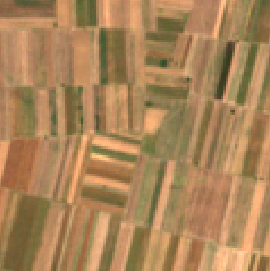}} 
    \end{minipage}    
          \begin{minipage}[t]{\figwidth}
        \centering
        \vspace{-0.47in}
        \centerline{\includegraphics[height=\figheight]{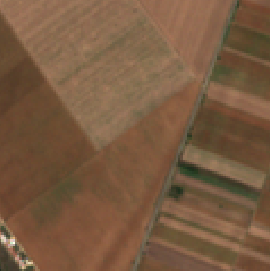}} 
    \end{minipage}
          \begin{minipage}[t]{\figwidth}
        \centering
        \vspace{-0.47in}
        \centerline{\includegraphics[height=\figheight]{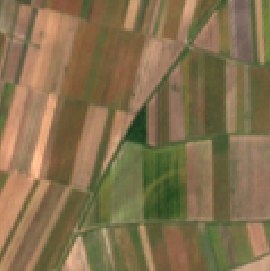}} 
    \end{minipage}
          \begin{minipage}[t]{\figwidth}
        \centering
        \vspace{-0.47in}
        \centerline{\includegraphics[height=\figheight]{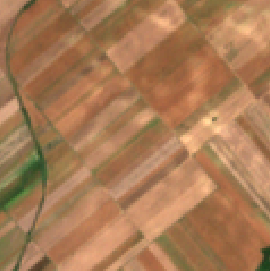}} 
        \vspace{-0.05in}\centerline{(c)}\medskip\vspace{-0.05in}
    \end{minipage}
          \begin{minipage}[t]{\figwidth}
        \centering
        \vspace{-0.47in}
        \centerline{\includegraphics[height=\figheight]{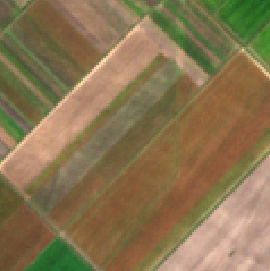}} 
    \end{minipage}
          \begin{minipage}[t]{\figwidth}
        \centering
        \vspace{-0.47in}
        \centerline{\includegraphics[height=\figheight]{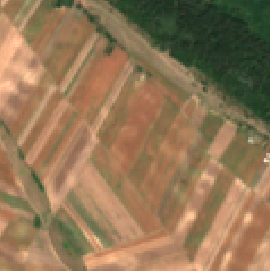}} 
    \end{minipage} 
          \begin{minipage}[t]{\figwidth}
        \centering
        \vspace{-0.47in}
        \centerline{\includegraphics[height=\figheight]{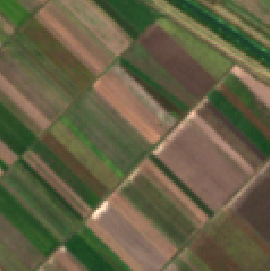}} 
    \end{minipage}
    \vspace{-0.3cm}
    \caption{(a) Query image; (b) images retrieved by the standard approach (which requires image decoding); and (c) images retrieved by our deep hashing based indexing module.}
    \label{fig:soa_comp_visres_figure_DLRSD}
    \vspace{-0.29cm}
\end{figure*}
Fig. 2 shows the PSNR results at different bits per pixel values obtained by our compression module and JPEG2000. By assessing the figure, one can observe that our compression module achieves the highest PSNR value at each bit-rate compared to JPEG2000, which is widely used in operational applications. This shows that the proposed framework is able to effectively decode RS images, when RS image compression and indexing are jointly learned. Table 1 reports the retrieval results obtained by the standard approach and our deep hashing based indexing module. One can see from the table that the retrieval accuracies obtained by our hashing based indexing module are very similar to those obtained by the standard approach. However, the required retrieval time of our indexing module is almost one-fourth of the time required for the standard approach. This is due to the fact that the retrieval time of standard approach includes the decoding time of all archive images, which is not required for our indexing module. In detail, since RS image compression and indexing are simultaneously learned by the proposed framework, hash codes generated by our indexing module are directly used for retrieval without decoding the data. Fig. 3 shows an example of images retrieved by the standard approach and our module. By analyzing the figure, one can observe that both approaches retrieve images semantically similar to the query image. This is in line with the results provided in Table 1.
\vspace{-0.05in}
\section{Conclusion}
\vspace{-0.05in}
In this paper, we have introduced a joint framework to simultaneously learn compression and indexing of RS images for accurate and fast CBIR. Our framework includes: i) a DL based compression module based on an auto-encoder architecture; and ii) a deep hashing based indexing module consisting of soft pairwise, bit-balancing and classification loss functions. We have also proposed a two stage learning strategy with gradient manipulation techniques to learn deep representations compatible with both compression and indexing. The effectiveness of our framework relies on accurate learning of quantized representation and hash codes of RS images at the same time. This leads to elimination of full decompression time, which is required for most of the indexing methods for RS CBIR. It is worth noting that the encoder of the proposed framework relies on the 2D convolutional layers, and thus gives equal importance to each image band. As a future work, we plan to integrate 3D convolutional layers to the encoder that can provide more accurate feature extraction in spectral domain. The use of 3D convolutions can lead to reducing the redundancy in image representations, and thus better rate-distortion trade-off for compression.
\vspace{-0.055in}
\section{Acknowledgements}
\vspace{-0.05in}
This work is funded by the European Research Council (ERC) through the ERC-2017-STG BigEarth Project under Grant 759764.
\bibliographystyle{IEEEbib}
\small 
\bibliography{defs, refs}
\end{document}